\documentclass[journal]{IEEEtran}

\usepackage{times}
\usepackage{epsfig}
\usepackage{graphicx}
\usepackage{amsmath}
\usepackage{amssymb}
\usepackage{multirow}
\usepackage{cite}
\usepackage{caption}
\usepackage{booktabs}
\usepackage{subfig}
\usepackage{makecell}
\usepackage{gensymb}
\usepackage{pifont}

\usepackage{color}
\definecolor{citecolor}{RGB}{119,185,0} 
\usepackage[pagebackref=true,breaklinks=true,letterpaper=true,colorlinks,citecolor=citecolor,bookmarks=false]{hyperref}
\usepackage{floatrow}
\usepackage{comment}

\usepackage[dvipsnames,svgnames,x11names]{xcolor}

\newcommand{\ie}{\mbox{\emph{i.e.}}}
\newcommand{\eg}{\mbox{\emph{e.g.}}}
\newlength\savewidth\newcommand\shline{\noalign{\global\savewidth\arrayrulewidth
  \global\arrayrulewidth 1pt}\hline\noalign{\global\arrayrulewidth\savewidth}}

\begin{document}
\title{Unsupervised Eyeglasses Removal in the Wild}

\author{Bingwen~Hu,
        Zhedong~Zheng,
        Ping~Liu,~\IEEEmembership{Member,~IEEE,}
        Wankou~Yang,~%~\IEEEmembership{Member,~IEEE,}
        and~Mingwu~Ren%~\IEEEmembership{Senior Member,~IEEE}% <-this  
        }


\maketitle

\begin{abstract}
Eyeglasses removal is challenging in removing different kinds of eyeglasses, \emph{e.g.,} rimless glasses, full-rim glasses and sunglasses, and recovering appropriate eyes. Due to the significant visual variants, the conventional methods lack scalability. Most existing works focus on the frontal face images in the controlled environment, such as the laboratory, and need to design specific systems for different eyeglass types. To address the limitation, we propose a unified eyeglass removal model called Eyeglasses Removal Generative Adversarial Network (ERGAN), which could handle different types of glasses in the wild. The proposed method does not depend on the dense annotation of eyeglasses location but benefits from the large-scale face images with weak annotations. Specifically, we study the two relevant tasks simultaneously, \emph{i.e.,} removing eyeglasses and wearing eyeglasses. Given two face images with and without eyeglasses, the proposed model learns to swap the eye area in two faces. The generation mechanism focuses on the eye area and invades the difficulty of generating a new face. In the experiment, we show the proposed method achieves a competitive removal quality in terms of realism and diversity. Furthermore, we evaluate ERGAN on several subsequent tasks, such as face verification and facial expression recognition. The experiment shows that our method could serve as a pre-processing method for these tasks.
%Our code is available at \url{https://github.com/Bingwen-Hu/ERGAN-Pytorch}.
\end{abstract}

% Note that keywords are not normally used for peer review papers.
\begin{IEEEkeywords}
Eyeglasses Removal, Image Manipulation, Generative Adversarial Network
\end{IEEEkeywords}

\IEEEpeerreviewmaketitle

\section{Introduction}
\IEEEPARstart{T}{he} eye is viewed as `` a window to the soul '' \cite{zebrowitz2018reading}, containing rich bio-metric information, \emph{e.g.,} identity, gender, and age. In recent years, there are increasing interests in face-related applications. 
Among these applications, eyeglasses are usually considered as one kind of occlusion in the face images. As a result, the occlusion compromises downstream tasks, such as face verification \cite{schroff2015facenet,mclaughlin2016largest,guo2017fuzzy,gaston2018matching,wang2019modal,guo2020learning,cao2020domain}, expression recognition~\cite{liu2014facial, liu2014feature, yang2018facial}. One way to address occlusion is by ignoring the occluded area, which successfully applied in the field of person re-identification~\cite{zhong2017random,zheng2018pedestrian,sun2018beyond,Sun_2019_CVPR,miao2019pose}. Different from the human body, the face is rich in identity information, and the eye area is the most discriminative facial field. The retained information is insufficient to support us make an accurate decision while the occluded area is ignored. Therefore, we propose an eyeglasses removal method to transform the occlusion area into a non-occlusion area. Despite significant advances in image manipulation, eyeglasses removal in the unconstrained environment, as known as in the wild, has not been well-studied. In this work, we intend to fill this gap.

\begin{figure}[t]
\vspace{+2mm}
\begin{center}
   \includegraphics[width=1\linewidth]{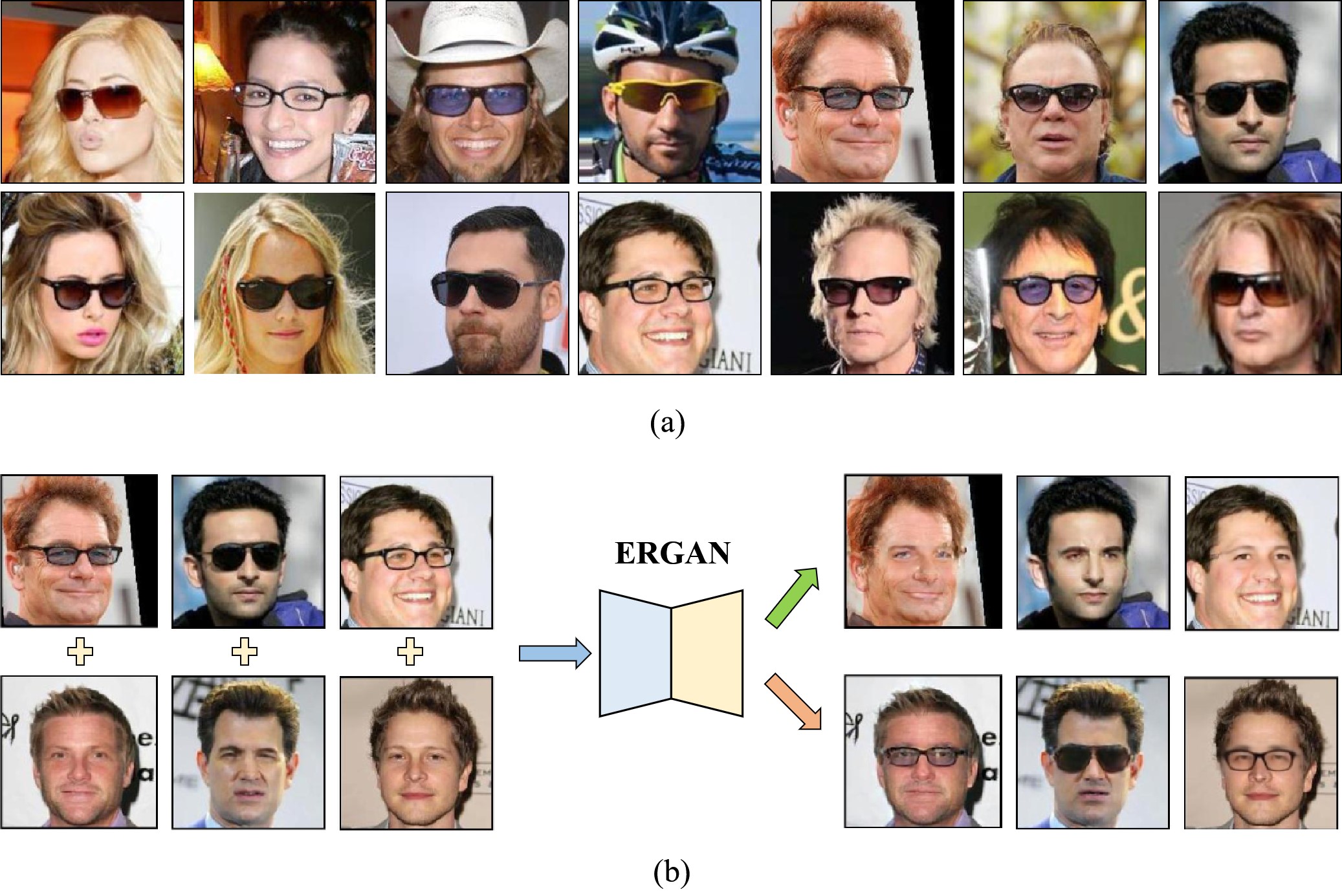}
\end{center}
\vspace{-.2in}
   \caption{(a) Examples of different types of eyeglasses in the wild. Different types of eyeglasses have significant visual variants, such as color, style, and transparency. Besides, the faces in the wild are usually in the arbitrary poses with various lighting and backgrounds. (b) A brief pipeline of the proposed method. The proposed Eyeglasses Removal Generative Adversarial Network (ERGAN) simultaneously takes the two relevant tasks, \emph{i.e.,} wearing and removing glasses, into consideration.}
\label{fig:glass}
\end{figure}

The previous eyeglasses removal works mainly focus on the cases in a controlled environment. Most works~\cite{wu2004automatic,park2005glasses,du2005eyeglasses,yi2011learning,wong2013eyeglasses} require pair-wise training samples.
Every pair of images contains two frontal faces of the same identity with and without glasses. When testing, given one frontal testing image, the framework first detects the eye area and then replaces the original eye area with the reconstructed eye. The crafted eye area fuses the no-glasses training samples by Principal Component Analysis (PCA)~\cite{pearson1901liii}. These lines of works adopt a strong assumption that faces are in a frontal pose, and the types of eyeglasses are limited. In the realistic scenario, however, there are three main drawbacks: \textbf{(1)} It is hard to collect a large number of pair-wise training data of the same people with and without eyeglasses; \textbf{(2)} Eyeglasses are usually made of different material with significant visual variants, such as color, style, and transparency (see Fig.~\ref{fig:glass} (a)). It is infeasible to train dedicated removal systems for different eyeglasses types; \textbf{(3)} Existing works could not be applied to the face images in different poses. The model trained in the laboratory environment usually lacks scalability to significant visual variants. 

This paper addresses these three challenges. First, the previous methods~\cite{wu2004automatic,park2005glasses,du2005eyeglasses,yi2011learning,wong2013eyeglasses} focus on the data collected in a controlled environment. More than that, they need the collected training images are pair-wised, \ie, two frontal faces of the same identity with and without glasses. Other than that, some works \cite{ wu2004automatic, park2005glasses,du2005eyeglasses} require an extra detector to locate the eyeglasses. Benefiting from the advancement of unsupervised learning~\cite{wei2017selective,wei2019unsupervised}, we propose an unsupervised eyeglasses removal method. The only information that we need is whether the training image contains eyeglasses on the face. For training the proposed model in this work, we have collected 202,599 training images from the public dataset CelebA~\cite{liu2015deep}. Those collected images are divided into two sets,~\ie, face images with glasses and face images without glasses, respectively. This weak labeling-demand for training sets largely saves the annotation cost.

% In particular, we collect the large-scale training images from public datasets on the web and divide them into two sets, \ie, the images with glasses and without glasses, respectively. The weak demand for training set largely saves the annotation cost.

Second, eyeglasses usually have significant intra-class variants in terms of geometry shape and appearance. It is infeasible to build dedicated models for every kind of eyeglasses. We, therefore, exploit an alternative method that lets the model ``see'' various eyeglasses and learn the scalable weights from a large number of face images. %Attribute to recent advances in neural networks, we acquire large-scale face images with various eyeglasses in the wild and deploy the Generative Adversarial Network (GAN). 
In particular, we propose the Eyeglass Removal GAN (ERGAN) to learn the general structure of glasses and encode different types of eyeglasses. The proposed method not only removes the eyeglasses but also has the capability to generate the eyeglasses, which further enforces the model to learn the local patterns of different eyeglasses (see Fig. \ref{fig:glass} (b)). 

Third, the problem of eyeglasses removal from arbitrary face poses is common and challenging in a realistic scenario. The previous works mostly assume that we could obtain frontal face images and accurately align the eye area. In real scenarios, the user, however, may upload non-frontal face images, and the eye-area is not well aligned. One solution is to resort to alignment calibration to generate frontal faces, which might be complicated and time-consuming. By contrast, the proposed method does not need alignment densely. The only pre-processing required is to rotate the face images according to the center of the face. Comparing to previous works, the proposed method does not need complicated alignment and is robust to the various pose variants.

% By training on the extensively collected face images which have arbitrary poses, the proposed method does not need complicated alignment and is robust to the various pose variants.

To address the challenges mentioned above, we propose a unified eyeglasses removal generative adversarial network (ERGAN), which only needs weak annotations for training. We learn two representations of the input face image, \ie, the face appearance code, and the eye-area attribute code. The face appearance code mainly contains the low-level geometric structure of the face, while the eye-area attribute code captures the semantic pattern in the eye area.
In more detail, we first utilize two different encoders to decompose the face image into a face appearance code and an eye-area attribute code, respectively. Then, one decoder is learned to combine the face appearance code with the eye-area attribute code to reconstruct the face image. The self-reconstruction loss is applied to ensure that the two latent codes are complementary to each other and preserve the information of the original image. To enforce the model focuses on the eye area of the input image, we introduce the eye-area reconstruction loss. Furthermore, the face appearance reconstruction loss and the cycle consistency loss are adopted to encourage that the mapping function is invertible between the reconstructed face image and the two latent codes. When testing, we could combine the face appearance code of the target face with the eye-area attribute code to remove or wear any specific eyeglasses.

To evaluate the generation quality, we adopt the FID \cite{heusel2017gans} and LPIPS \cite{zhang2018unreasonable} as an indicator to test the realism and diversity of the generated face images, respectively. 
Extensive qualitative and quantitative experiments show that our method achieves superior performance to several existing approaches~\cite{CycleGAN2017,liu2017unsupervised,huang2018munit} and could serve as a pre-processing tool for subsequent tasks, such as face verification and facial expression recognition. 
%Compared to these existing approaches~\cite{CycleGAN2017,liu2017unsupervised,huang2018munit} to manipulate the whole face image, our proposed method focuses on the manipulation of the eye-area, which avoids interference of irrelevant information.
%Given a pair of face images (with and without glasses), swap the  eye-area attribute codes from both face images, and then the swapped  eye-area attribute codes are combined with the original face appearance codes to obtain the results of removing glasses and wearing glasses respectively.
% and the image in each domain has a different distribution. The two domains with different distribution can be translated by using the generative adversarial network. Fig \ref{fig:ERGAN} shows the detailed illustration of our method. We build our method on the the success of recent image-to-image translation frameworks such as~\cite{tran2017disentangled,liu2017unsupervised,huang2018munit,lee2018diverse}. 
 %To be specific, we present a unified eyeglasses removal framework  to remove different types of eyeglasses in the unconstrained condition. 

In summary, the main contributions of this work are as follows:  
\begin{itemize}
    \item We propose a unified framework called ERGAN that enables different types of eyeglasses removal from faces in the wild. Compared with previous works, the proposed method does not need densely-aligned faces nor pair-wise training samples. The only weak annotation that we need is whether the training data contains eyeglasses or not. 
    \item Due to the large visual variants of the eyeglasses, including shape and color, eyeglasses removal demands to learn a robust model. In this work, we propose a dual learning scheme to simultaneously learn two inverse manipulations, \ie, removing eyeglasses, and wearing eyeglasses. Specifically, we utilize the eye-area reconstruction loss to explicitly make the model pay more attention to the eye area. The ablation study verifies the effectiveness of both the dual learning scheme and the eye-area reconstruction loss. 
    \item The qualitative and quantitative experiments show that our method outperforms other competitive approaches in terms of realism and diversity. Furthermore, we evaluate the proposed method on other face-related tasks. Attribute to the high-fidelity eyeglass removal, the generated results benefit the subsequent tasks, such as face verification and facial expression recognition. 
\end{itemize}

\section{Related work}
\subsection{Statistical Learning}
Most pioneering works on eyeglasses removal are based on statistical learning~\cite{wu2004automatic,park2005glasses,de2006generalized,yi2011learning,wong2013eyeglasses}. One line of works adopts the assumption that the target faces could be reconstructed from other faces without eyeglasses. Based on this assumption, previous methods widely adopt Principal Component Analysis (PCA) \cite{pearson1901liii} to learn the shared components among the face images. In one of the early works, Wu \emph{et al.} \cite{wu2004automatic} proposes a find-and-replace approach to remove eyeglasses from frontal face images. This method first finds the location of eyeglasses by an eye-region detector and then replaces it with a synthesized glasses-free image. The synthesized glass-free image is inferred by combining the original eye area and different weighted glasses-free faces in the training data. Furthermore, Park \emph{et al.} \cite{park2005glasses} apply the recursive process of PCA reconstruction and error compensation to generate facial images without glasses. 
Due to the different temperatures between glasses and human faces, another line of works takes advantage of thermal images to remove the eyeglasses. Wong \emph{et al.} \cite{wong2013eyeglasses} proposes a nonlinear eyeglasses removing algorithm for thermal images based on kernel principal component analysis (KPCA)~\cite{mika1999kernel}. This method performs KPCA to transfer the visible reconstruction information from the visible feature space to the thermal feature space, and then apply the image reconstruction to remove eyeglasses from the thermal face image.
Different from the method based on PCA mentioned above, some researchers resort to sparse coding and expectation-maximization to reconstruct faces. Yi \emph{et al.}~\cite{yi2011learning} deploys the sparse representation technique in local feature space to deal with the issue of eyeglasses occlusion. De  Smet \emph{et al.}~\cite{de2006generalized} proposed a generalized expectation-maximization approach, which applies the visibility map to inpaint the occluded areas of the face image. 

However, these methods \cite{wu2004automatic,park2005glasses,de2006generalized,yi2011learning,wong2013eyeglasses} are usually designed for frontal faces and specific eyeglasses in a controlled environment. Different from the existing work, our method focuses on face images collected from the realistic scenario, and is scalable to visual variants, such as pose, illumination, and different types of glasses.
\vspace{-.15in}
\subsection{Generative Adversarial Network}
Recent advance in facial manipulation is due to two factors: \textbf{(1)} large-scale public face datasets with attribute annotations, \eg, CelebA \cite{liu2015deep}; \textbf{(2)} The high-fidelity images generated by Generative Adversarial Network (GAN). GAN is one kind of the generative model benefiting from the competition between the generator and the discriminator \cite{goodfellow2014generative,zhong2018camstyle,yang2019very,luo2019significance,zhu2019dm,luo2020adversarial}. The facial image manipulation algorithms based on GANs have taken significant steps \cite{shen2017learning,CycleGAN2017,liu2017unsupervised,huang2018munit,zhang2018generative,liu2019stgan}. 
%Face attribute editing based on GANs is usually treated as an unpaired image-to-image translation task . 
One line of previous work directly learns the image-to-image translation between different facial attributes. Shen \emph{et al.} \cite{shen2017learning} presents a GAN-based method using residual image learning and dual learning to manipulate the attribute-specific face area. Liu \emph{et al.} \cite{liu2019stgan} presents a unified selective transfer network for arbitrary image attribute editing (STGAN), by combining difference attribute vector and selective transfer unit (STUs) in autoencoder network. 
Another line of works first learn the embedding of the face attributes and then decode the learned feature to generate images.
Liu \emph{et al.} \cite{liu2017unsupervised} proposes an unsupervised image-to-image translation (UNIT) framework, which combines the spirit of both GANs and variational autoencoders (VAEs) \cite{kingma2013auto}. UNIT adopts the assumption that a pair of images in different domains can be mapped to a shared latent feature space, and thus, could conduct the face images translation by manipulating the latent code. Furthermore, Huang \emph{et al.} proposes a method for the multimodal unsupervised image-to-image translation (MUNIT) \cite{huang2018munit}. MUNIT assumes that a pair of images in different domains share the same content space but the style space. Sampling different style codes could produce diverse and multimodal outputs while preserving the principle content. 

\begin{figure*}[t]
%\vspace{+2mm}
\begin{center}
   \includegraphics[width=1\linewidth]{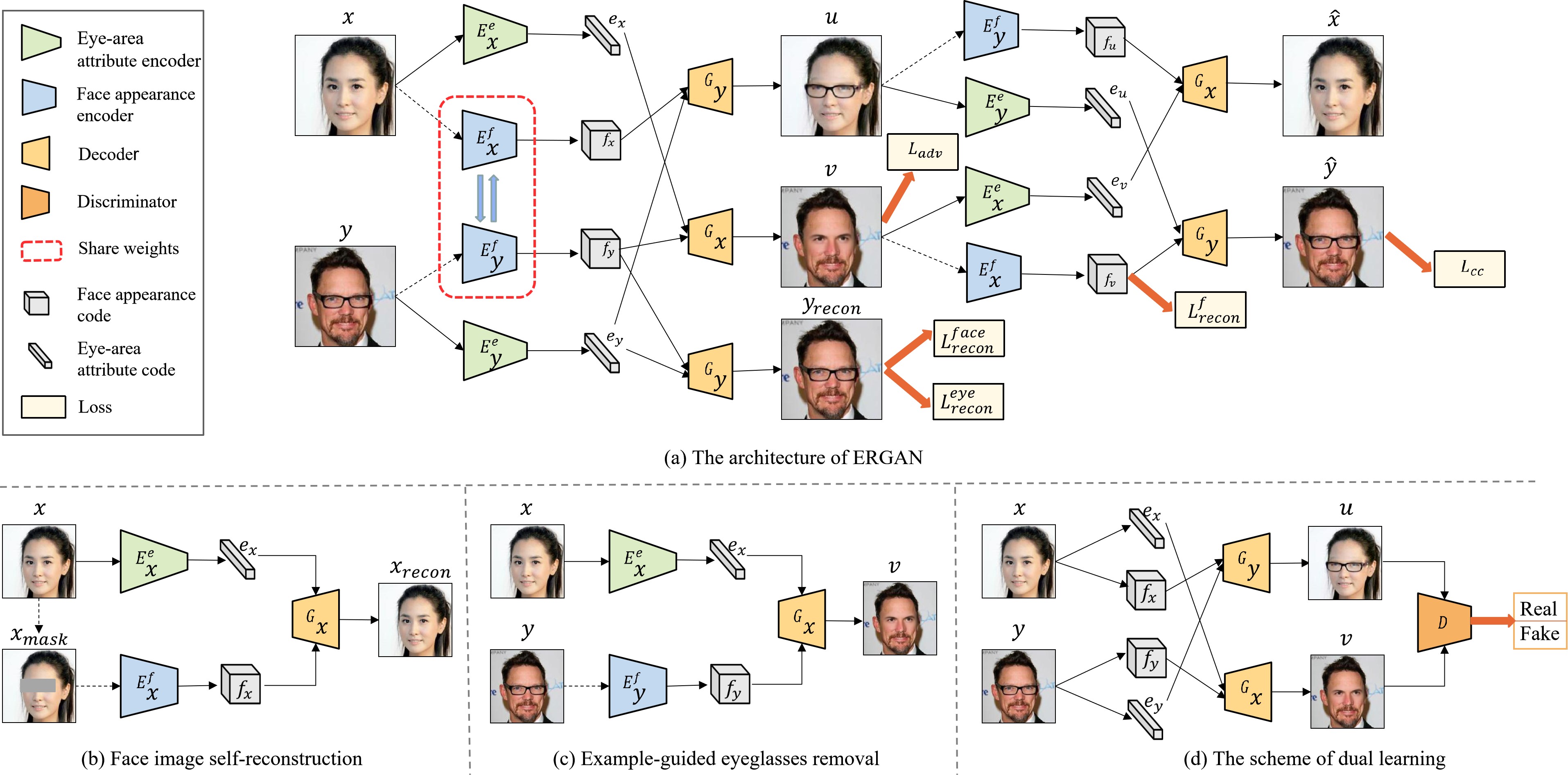}
   \vspace{-.2in}
\end{center}
   \caption{An overview of Eyeglasses Removal Generative Adversarial Network (ERGAN). 
   (a) The proposed method implements two mappings: $ \mathcal{X} \rightarrow{} \mathcal{Y} $ and $ \mathcal{Y} \rightarrow{} \mathcal{X} $. The input face image is decomposed to a face appearance code $f$ and an eye-area attribute code $e$ by encoders $E^{f}_{.}$ and  $E^{e}_{.}$, respectively. Encoders $E^{f}_\mathcal{X}$ and $E^{f}_\mathcal{Y}$ share weights. The black dash line denotes that the eye region of the input image to face appearance encoder $E^{f}_{.}$ is masked. 
   (b) The self-reconstruction process of input image $x$ to generate $x_{recon}$. $x_{mask}$ denotes $x$ after the eye area is masked.
   (c) The illustration of example-guided eyeglasses removal. $e_{x}$ and $f_{y}$ are combined to generate $v$ by $G_\mathcal{X}$. 
   (d) We propose a dual learning scheme to simultaneously learn two inverse manipulations (removing eyeglasses and wearing eyeglasses) by swapping eye-area attribute codes.}
\label{fig:ERGAN}

\end{figure*}

% In this work, we also adopt the GAN-based network, called ERGAN, but we are different from these GAN-based approaches~\cite{CycleGAN2017,liu2017unsupervised,huang2018munit} in the following main aspects: 
% Comparing to previous GAN-based approaches ~\cite{CycleGAN2017,liu2017unsupervised,huang2018munit}, the proposed ERGAN has significant differences listed as follows: \textbf{(1)} Attribute to the eye-area loss and invertible eye generation, the proposed ERGAN explicitly focuses on the manipulation of the eye region, while the existing works focus on generating the whole face. The conventional generation mechanism inevitably introduces the noise to other areas of the original face when removing glasses; \textbf{(2)} We adopt the instance generation mechanism, which could swap the eye area of any two facial images. Compared with the CycleGAN-based methods, we could generate more diverse images; \textbf{(3)} Different from the previous works treat the face with different attributes as two or multiple domains, we view the face images as one domain with two codes, \ie, the face appearance  code, and  the eye-area attribute code. We could further manipulate the code to generate conditional outputs to meet user demands.  %We share the weights of the encoders, which saves the model parameters and improves eyeglass removal performance.
%Although the aforementioned GANs-based facial images manipulation algorithm has been significantly improved compared with the statistical-learning-based method, there are still some shortcomings, 
Comparing to previous GAN-based approaches ~\cite{CycleGAN2017,liu2017unsupervised,huang2018munit}, the proposed ERGAN has significant differences listed as follows: \textbf{(1)} To the introduced eye-area loss and invertible eye generation, the proposed ERGAN explicitly focuses on the manipulation of the eye region. The conventional generation mechanism inevitably introduces the noise to other areas of the original face when removing glasses; \textbf{(2)} We adopt the instance generation mechanism, which could swap the eye area of any two facial images. Compared with the CycleGAN-based methods, the proposed method could generate more diverse images; \textbf{(3)} Different from the previous works treating the face with different attributes as two or multiple domains, we view the face images as one domain with two codes, \ie, the face appearance code, and the eye-area attribute code. By utilizing this fine-grained latent representation decomposition, we could manipulate the code to generate conditional outputs to meet user demands. 

%\IEEEpubidadjcol

\section{Methodology}
We first formulate the problem of eyeglasses removal and provide an overview of the proposed eyeglasses removal generative adversarial network (ERGAN) in Section~\ref{3.0}. In Section \ref{3.2}, we describe the details of each component in ERGAN, followed by the full objective function in Section \ref{3.3}.

\subsection{Formulation} 
\label{3.0}
In this work, we assume each face image could be decomposed into a face appearance code and an eye-area attribute code. Bases on the assumption, we combine a face appearance code with an eye-area attribute code from an example face image to remove or wear eyeglasses. In this case, we could handle two inverse manipulations at the same time, \emph{i.e.,} eyeglasses removal and eyeglasses wearing.
The architecture of the proposed unified eyeglasses removal generative adversarial network (ERGAN) is presented in Fig. \ref{fig:ERGAN} (a). We denote face images with glasses and without glasses as $ \mathcal{X} $ and $ \mathcal{Y} $ ( $ \mathcal{X}, \mathcal{Y} \subset \mathbb{R}^{H \times W \times 3} $ ), respectively. The goal of the proposed method is to learn two mappings: $ \mathcal{X} \rightarrow  \mathcal{Y} $ and $ \mathcal{Y} \rightarrow \mathcal{X}$ that could transfer an image $x \in \mathcal{X}$ to another image $y \in \mathcal{Y}$, and vice versa. 

\textbf{Generator.} As illustrated in Fig. \ref{fig:ERGAN} (a), the generator of the proposed method adopts a similar framework of auto-encoder~\cite{hinton1994autoencoders,zhu2017uncovering}, which consists of face appearance encoders $ (E^{f}_\mathcal{X}, E^{f}_\mathcal{Y})$, eye-area attribute encoders $ (E^{e}_\mathcal{X}, E^{e}_\mathcal{Y}) $, and decoders $ (G_\mathcal{X}, G_\mathcal{Y})$. Specifically, the role of encoders $ (E^{f}_{.}, E^{e}_{.}) $ is to encode a given face image into the face appearance code $f_{.}$ and the eye-area attribute code $e_{.}$, respectively. Moreover, $f_{.}$ and $e_{.}$ are combined to generate a new image with the decoder $ G_{.}$. The decoder $G_{.}$ is a deterministic function and has inverse encoders $ (E^{f}_{.}, E^{e}_{.}) = (G_{.})^{-1} $. In particular, $G_\mathcal{X}$ generates face images without glasses and $G_\mathcal{Y}$ generates face images with glasses, respectively. 

\textbf{Discriminator.} $ (D_\mathcal{X}, D_\mathcal{Y}) $ are two discriminators for $ \mathcal{X} $ and $ \mathcal{Y} $, respectively. The discriminator $D_{.}$ aims to distinguish between generated images and real images. For instance, given a pair of images $x \in \mathcal{X}$ and $y \in \mathcal{Y}$, the discriminator $D_\mathcal{X}$ aims to distinguish images generated by decoder $G_\mathcal{X}(E^{f}_\mathcal{X}(x),E^{e}_\mathcal{Y}(y))$ from real images in $\mathcal{X}$. In the same way, the discriminator $D_\mathcal{Y}$ aims to distinguish images generated by decoder $G_\mathcal{Y}(E^{f}_\mathcal{Y}(x),E^{e}_\mathcal{X}(x))$ from real images in $\mathcal{Y}$. Specifically, the generated image by decoder $G_\mathcal{X}(E^{f}_\mathcal{X}(x),E^{e}_\mathcal{Y}(y))$ has the same face appearance code of $x$ and the same eye-area attribute code of $y$. By contrast, the generated image by decoder $G_\mathcal{Y}(E^{f}_\mathcal{Y}(x),E^{e}_\mathcal{X}(x))$ has the same face appearance code of $y$ and the same eye-area attribute code of $x$.
\vspace{-.15in}
\subsection{Eyeglasses Removal Generative Adversarial Network}
\label{3.2}
\textbf{Face image self-reconstruction.} As shown in Fig. \ref{fig:ERGAN} (b), to enforce the generator focuses on the eye-area of the input face image $x$, we first mask out the eye region to generate $x_{mask}$ and then encode $x$ and $x_{mask}$ by encoders $(E^{f}_\mathcal{X}, E^{e}_\mathcal{X})$ to obtain the eye-area attribute code $e_{x}$ and the face appearance code $f_{x}$. Finally, $e_{x}$ and $f_{x}$ are combined to generate the self-reconstructed image $x_{recon}$ with the decoder $G_\mathcal{X}$, where $x_{recon}=G_\mathcal{X}(E^{f}_\mathcal{X}(x),E^{e}_\mathcal{X}(x))$. It is straight-forward that the generated result of self-reconstruction approximates the source image. We introduce the face self-reconstruction loss, which is defined as:

\begin{align}
\begin{split}
     L^{face}_{recon}=\mathbb{E}[\|G_\mathcal{X}(E^{f}_\mathcal{X}(x),E^{e}_\mathcal{X}(x))-x\|_{1}]+\\
     \mathbb{E}[\|G_\mathcal{Y}(E^{f}_\mathcal{Y}(y),E^{e}_\mathcal{Y}(y))-y\|_{1}],
\end{split}
\label{eq1}
\end{align}
where $\mathcal{X}$ represents the set of face images without glasses and $\mathcal{Y}$ represents the set of face images with glasses, $x\in \mathcal{X}$ and $y\in \mathcal{Y}$. The pixel-wise $\ell_{1}$-norm $\|\cdot\|_{1}$ is employed for preserving the sharpness of self-reconstruction images. The face self-reconstruction loss enforces the encoders $(E^{f}_\mathcal{X}, E^{e}_\mathcal{X})$ to learn two representations of the input face image, \emph{i.e.,} the face appearance code and the eye-area attribute code, respectively. 

\textbf{Eye-area reconstruction.}
The face image self-reconstruction loss encourages the model to focus on the global features of the input image. For the glasses removal task, employing only the face image self-reconstruction loss misses the specific information of the eye area of the generated image. Therefore, enforcing the model to pay more attention to the eye area, we introduce the eye-area reconstruction loss:
%\vspace{-0.2mm}
\begin{align}
    L^{eye}_{recon} = \mathbb{E}[\|x^{eye}_{recon}-x^{eye}\|_{1}]+
    \mathbb{E}[\|y^{eye}_{recon}-y^{eye}\|_{1}],
\end{align}
where $x^{eye}$ and $x^{eye}_{recon}$ are denoted the eye-area of $x$ and $x_{recon}$, respectively. Similarly, $y^{eye}$ and $y^{eye}_{recon}$ are denoted the eye-area of $y$ and $y_{recon}$.
In practice, since the face images are all center-aligned, the eye area is defined as $(x_1 = 0.4h, x_2 = 0.2w, y_1 = 0.65h, y_2 = 0.75w)$ area of the input image, where $(h, w)$ is the size of the input face image. 

\textbf{Dual learning scheme.} The proposed method is based on the assumption that the face image could be decomposed into a face appearance code and an eye-area attribute code. In detail, given a face appearance code, the decoder $G_{.}$ could combine the face appearance code with the eye-area attribute code from the target face image to generate an image with or without glasses. Fig. \ref{fig:ERGAN} (c) shows an example of eyeglasses removal. The two related tasks of removing glasses and wearing glasses could be regarded as a dual process. Therefore, we apply a dual learning scheme~\cite{yi2017dualgan} to learn two inverse manipulations simultaneously. The ablation study (in Section \ref{ablation}) confirms the effectiveness of the dual learning scheme. 

We show the process of dual learning in Fig. \ref{fig:ERGAN} (d). For given images $x$ and $y$, we encode them into $(f_{x},e_{x})$ and $ (f_{y},e_{y})$, where $(f_{x},e_{x})= (E^{f}_\mathcal{X},E^{e}_\mathcal{X})$ and $(f_{y},e_{y})= (E^{f}_\mathcal{Y},E^{e}_\mathcal{Y})$. The dual tasks (\ie, removing eyeglasses and wearing eyeglasses) is performed by swapping the eye-area attribute codes. We adopt decoders $G_\mathcal{X}$ and $G_\mathcal{Y}$ to generate the final output images $u=G_\mathcal{Y}(f_{x},e_{y})$ and $v=G_\mathcal{X}(f_{y},e_{x})$, respectively. In particular, the decoder $G_\mathcal{Y}$ combines the face appearance code of $x$ and the eye-area attribute code of $y$ to generate $u$. Similarly, the decoder $G_\mathcal{X}$ combines the face appearance code of $y$ and the eye-area attribute code of $x$ to generate $v$. 

Based on the assumption that each face image could be decomposed into a face appearance code and an eye-area attribute code, in this work, we learn two representations of the input face image, \ie, the face appearance code, and the eye-area attribute code. 
The face appearance code mainly contains the low-level geometric structure of the face. Since the encoder maps the region outside the eye area to the face appearance space is irrelevant to the face image with or without glasses. 
During the training, therefore, we employ the weights sharing between $E^{f}_\mathcal{X}$ and $E^{f}_\mathcal{Y}$ to update the model synchronously. By contrast, the eye-area attribute code captures the semantic pattern in the eye area, resulting in the eye-area attributes of face images with glasses or without glasses are significantly different. In this case, we do not share weights between $E^{e}_\mathcal{X}$ and $E^{e}_\mathcal{Y}$.

Furthermore, the proposed method should be able to reconstruct $(f_{x},e_{x})$ and $(f_{x},e_{y})$ after decoding $u$ and $v$. As illustrated in Fig. \ref{fig:ERGAN} (d), we apply encoders $E^{f}_\mathcal{X}$ and $E^{e}_\mathcal{Y}$ obtain the face appearance code $f_{x}$ and the eye-area attribute code $e_{y}$, then $f_{x}$ and $e_{y}$ are concatenated together to generate $u$. A similar process is utilized to generate $v$. Then we encode $u$ and $v$ into $(f_{u},e_{u})$ and $(f_{v},e_{v})$.
To ensure that generated images $u$ and $v$ retain the original information, we define the face appearance reconstruction loss $L^{f}_{recon}$ and the eye-area attribute reconstruction loss $L^{e}_{recon}$. We formulate $L^{f}_{recon}$ and $L^{e}_{recon}$ as follow:

\begin{align}
    L^{f}_{recon}= \mathbb{E}[\|f_{u}-f_{x}\|_{1}]+\mathbb{E}[\|f_{v}-f_{y}\|_{1}]\\
    L^{e}_{recon}= \mathbb{E}[\|e_{u}-e_{x}\|_{1}]+\mathbb{E}[\|e_{v}-e_{y}\|_{1}],
\label{eq5}
\end{align}
where $(f_{u},f_{v})=(E^{f}_\mathcal{X}(u),E^{f}_\mathcal{Y}(v))$ and $(e_{u},e_{v})=(E^{e}_\mathcal{X}(u),E^{e}_\mathcal{Y}(v))$.
Notably, we find that the performance of the model is declined by introducing the eye-area attribute reconstruction loss $L^{e}_{recon}$. To achieve the highest performance, we ignore $L^{e}_{recon}$. The detailed discussion in our ablation study (in Section \ref{ablation}).

Then we recombine face appearance codes $(f_{u},f_{v})$ and eye-area attribute codes $(e_{u},e_{v})$ to generate $\hat{x}=G_\mathcal{X}(f_{u},e_{v})=G_\mathcal{X}(E^{f}_\mathcal{X}(u),E^{e}_\mathcal{Y}(v))$ and $\hat{y}=G_\mathcal{Y}(f_{v},e_{u})=G_\mathcal{Y}(E^{f}_\mathcal{Y}(v),E^{e}_\mathcal{X}(u))$, respectively. To ensure the generated images $u$ and $v$ reconstruct the origin images $x$ and $y$, we introduce the cycle consistency loss~\cite{CycleGAN2017}. The cycle consistency loss $L_{cc}$ is defined as:

\begin{align}
\begin{split}
    L_{cc} = \mathbb{E}[\|G_\mathcal{X}(E^{f}_\mathcal{X}(u),E^{e}_\mathcal{Y}(v))-x\|_{1}]+\\
    \mathbb{E}[\|G_\mathcal{Y}(E^{f}_\mathcal{Y}(v),E^{e}_\mathcal{X}(u))-y\|_{1}],
\end{split}
\end{align}
where $u=G_\mathcal{Y}(f_{x}, e_{y})$ and $v=G_\mathcal{X}(f_{y}, e_{x})$.

\textbf{Adversarial loss.} To encourage the generated face image indistinguishable from the real face image, we adopt the adversarial loss~\cite{goodfellow2014generative}. In this case, $G_\mathcal{X}$ and $G_\mathcal{Y}$ attempt to generate high-fidelity face images (\emph{e.g.}, with glasses and without glasses), while $D_\mathcal{X}$ and $D_\mathcal{Y}$ attempt to distinguish real face images from generated face images. The adversarial loss is defined as:

\begin{align}
\begin{split}
    L_{adv} = \mathbb{E}[\log D_\mathcal{X}(x)] + \mathbb{E}[\log(1-D_\mathcal{X}(G_\mathcal{X}(f_{y},e_{x})))]+\\
    \mathbb{E}[\log D_\mathcal{Y}(y)] + \mathbb{E}[\log(1-D_\mathcal{Y}(G_\mathcal{Y}(f_{x},e_{y})))].
\end{split}
\end{align}

\textbf{Discussion.} Different from several image-to-image translation frameworks~\cite{CycleGAN2017,liu2017unsupervised,huang2018munit} to manipulate the whole image, the proposed method focuses on the manipulation of the eye area, which does not change the regions outside the eye area and preserves the information of the original image effectively. Specifically, inspired by image inpainting works~\cite{bertalmio2000image,pathak2016context, yu2019free}, we cover the eye-area of the input face image and then apply the encoder $E^{f}_{.}$ to obtain the face appearance code, which avoids the interference of original eye area information. Moreover, we use the encoder $E^{e}_{.}$ to obtain the eye-area attribute code of the input face image. In particular, introducing both the face image self-reconstruction loss and the eye-area reconstruction loss enforces the model to learn two representations of the input image, \emph{i.e.,} the face appearance code and the eye-area attribute code. Besides, both the face image self-reconstruction loss and the eye-area reconstruction loss enforces the proposed model only to manipulate the eye region while maintaining the rest regions unchanged. 
Furthermore, we argue that the information feedback mechanism by dual learning could effectively improve the performance of the proposed method. Accordingly, we apply the dual learning scheme to realize the two inverse tasks of wearing glasses and removing glasses. The ablation study (in Section \ref{ablation}) demonstrate the effectiveness of the dual learning scheme.

\begin{figure*}[t]
\begin{center}
   \includegraphics[width=1.0\linewidth]{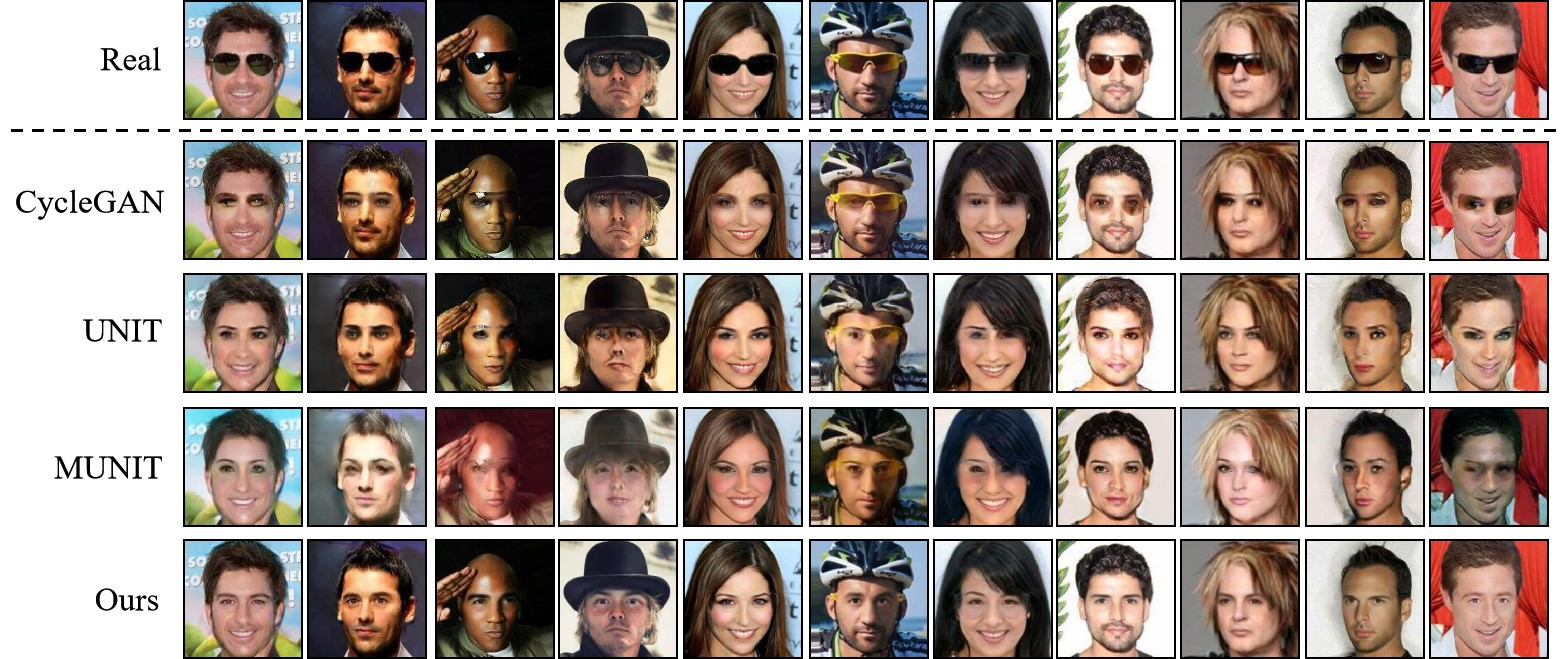}
 \end{center}
   \caption{Results of eyeglasses removal from face images on the CelebA dataset. From top to bottom: real images, CycleGAN~\cite{CycleGAN2017}, UNIT~\cite{liu2017unsupervised}, MUNIT~\cite{huang2018munit} and our method.}
 \label{fig:compare}
 \end{figure*} 
 
\begin{figure*}[t]
\begin{center}
   \includegraphics[width=1.0\linewidth]{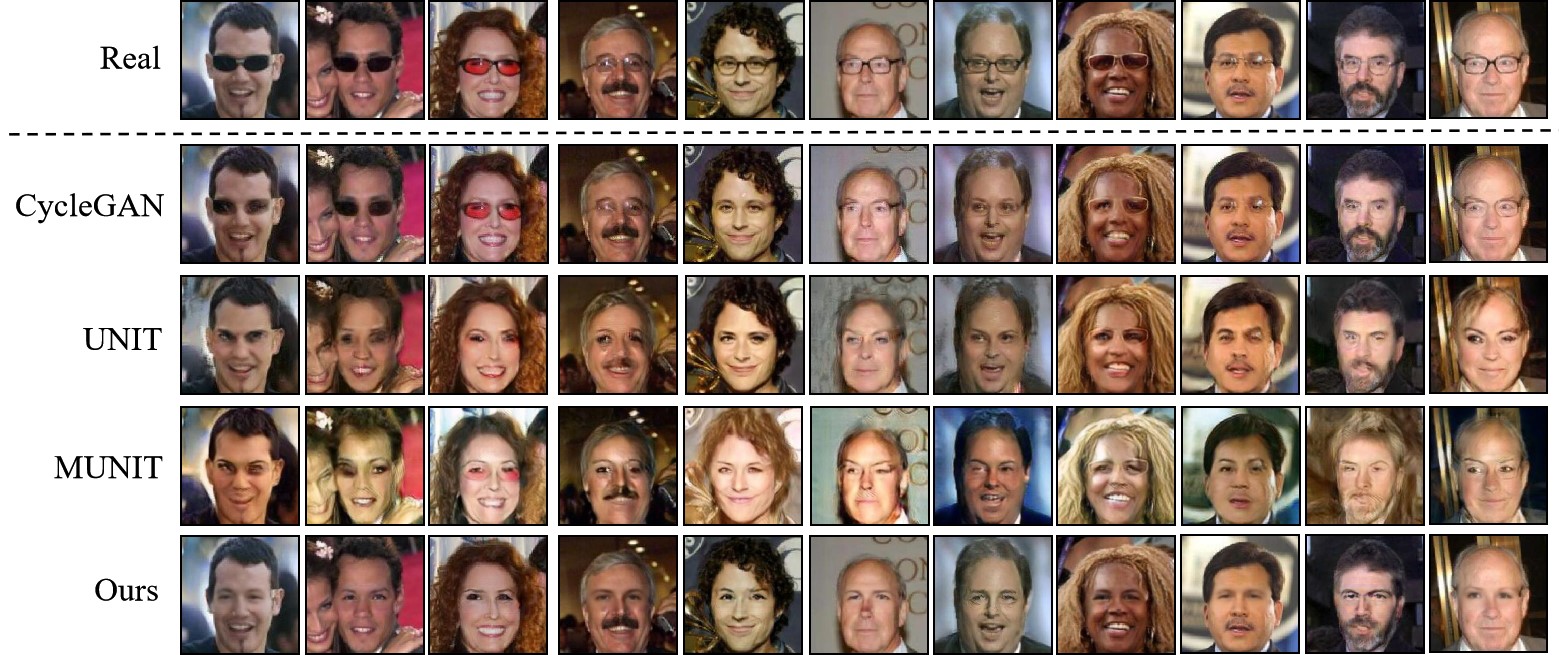}
 \end{center}
   \caption{Results of eyeglasses removal from face images on the LFW dataset. From top to bottom: real images, CycleGAN~\cite{CycleGAN2017}, UNIT~\cite{liu2017unsupervised}, MUNIT~\cite{huang2018munit} and our method.}
 \label{fig:compare_lfw}
 \end{figure*} 
 
\vspace{-.15in}
\subsection{Objective Function}
\label{3.3}
Taking all above loss functions into account, we jointly train the encoders, decoders, and discriminators. We formulate the full objective function as:

\begin{align} 
\begin{split}
\underset{E^{e}_{.},E^{f}_{.},G_\mathcal{X},G_\mathcal{Y}}\min\ \underset{D_\mathcal{X},D_\mathcal{Y}}\max\ L_{total}(E^{e}_{.},E^{f}_{.},G_\mathcal{X},G_\mathcal{Y},D_\mathcal{X},D_\mathcal{Y})=\\
\lambda_{face}L^{face}_{recon}+ \lambda_{eye}L^{eye}_{recon}+L^{f}_{recon}+L_{cc}+L_{adv},
\end{split}
\label{loss}
\end{align}
where the hyper-parameters $\lambda_{face}$ and $\lambda_{eye}$ control the weights of the face self-reconstruction loss and the eye-area reconstruction loss.
 
\section{Experiment}

\subsection{Datasets}

\textbf{CelebA.} The CelebA dataset \cite{liu2015deep} consists of 202,599 aligned face images collected from 10,177 celebrities in the wild. Each image in CelebA is annotated with 40 face attributes (\emph{e.g., with/without eyeglasses, smiling/no-smiling}). We split the CelebA dataset into one subset with glasses and another without glasses, based on the annotated attributes. Accordingly, we obtain 13,193 images with glasses and 189,406 images without glasses. All face images are center-cropped to $160\times160$ and resized to $224\times224$ with a probability of $0.5$ horizontal flipping.

\textbf{LFW.} The LFW dataset \cite{LFWTech} contains 13,233 face images of 5,749 identities collected from the uncontrolled surroundings. All images are aligned by deep funneling \cite{Huang2012a}. We choose 1,600 face images with glasses and 8,000 face images without glasses from the LFW dataset to verify the effectiveness of the proposed method. The preprocessing of the LFW dataset is the same as the CelebA dataset.

\textbf{MeGlass.} The MeGlass dataset~\cite{guo2018face} includes 47,917 face images of 1,710 identities selected and cleaned from the MegaFace dataset~\cite{kemelmacher2016megaface}. The MeGlass dataset contains 14,832 face images with glasses and 33,087 face images without glasses. Besides, Each identity contains at least two face images with glasses and two face images without glasses. We also perform experiments on MeGlass to verify the effectiveness of our method.
\vspace{-.2in}
\subsection{Evaluation Metrics}
\textbf{Fr\'{e}chet Inception Distance (FID).} Most image generation tasks~\cite{bulat2018learn,zheng2019joint,mao2019mode,razavi2019generating} adopt the FID metric \cite{heusel2017gans}, which is utilized to measure the distance between generated images and real images through feature extracted by Inception Network~\cite{szegedy2015going}. In this work, we apply the FID metric to measure the realism of the generated face images. 

\textbf{Learned Perceptual Image Patch Similarity (LPIPS).} The LPIPS \cite{zhang2018unreasonable} distance is used to evaluate the diversity of generated images. Following several state-of-the-art image-to-image translation approaches~\cite{zhu2017toward,huang2018munit,lee2018diverse,mao2019mode}, we employ the LPIPS distance to measure the diversity of manipulated face images. Because the proposed method focuses on the task of eyeglasses removal, we only capture the eye area to measure the LPIPS distance to make a fair comparison. We denote the modified LPIPS metric as~\textbf{eLPIPS}. 
% Note that the proposed method focuses on the task of eyeglasses removal. Therefore, to fairly compare the diversity of the generated images from different methods, we only capture the eye area to measure the LPIPS distance. We denote the modified LPIPS metric as~\textbf{eLPIPS}.

\begin{figure}[t]
\begin{center}
   \includegraphics[width=1.0\linewidth]{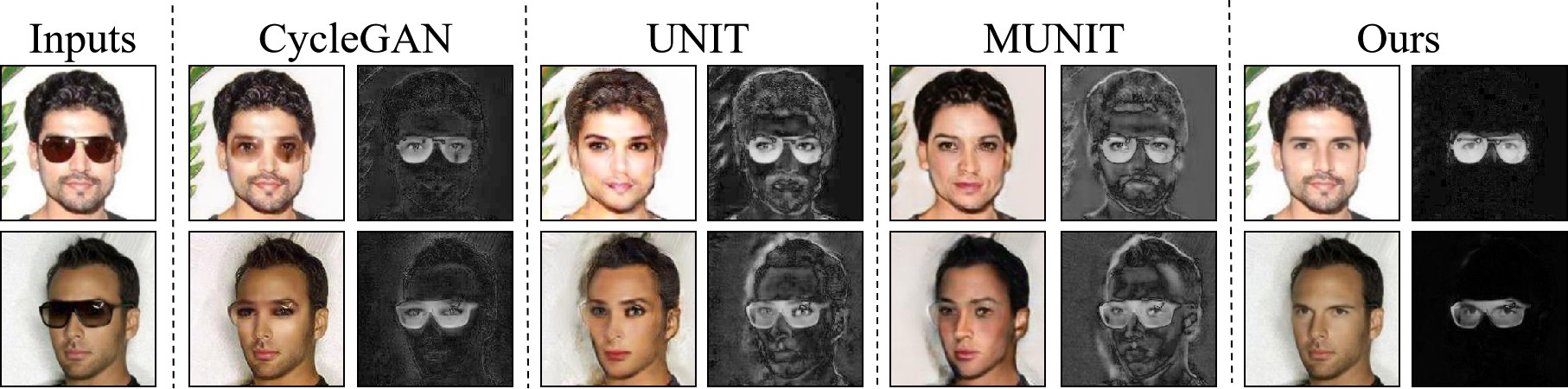}
 \end{center}
   \caption{The residual images obtained by our method and three competitive methods, \emph{e.g.}, CycleGAN~\cite{CycleGAN2017}, UNIT~\cite{liu2017unsupervised}, and MUNIT~\cite{huang2018munit}. }
 \label{fig:residual}
 \end{figure}
 
\vspace{-.15in}
\subsection{Implementation Details}
%\vspace{-0.5in}
Here we provide details about the network architecture of ERGAN.
\textbf{(1)} The eye-area attribute encoder $E^{e}_{.}$ consists of several convolutional layers and residual blocks ~\cite{he2016deep} as well as a global average pooling layer and a fully connected layer. \textbf{(2)} For the face appearance encoder $E^{f}_{.}$, it only consists of several convolutional layers and residual blocks. \textbf{(3)} The decoder $G_{.}$ combines $E^{e}_{.}$ and $E^{f}_{.}$ through four residual blocks followed by several convolutional layers and upsampling layers. In particular, we finally append a refine block which consists of a convolutional layer and a fully connected layer. The refine block is used to concatenate the reconstructed image with the input image to produce a higher quality generated image. Additionally, all convolutional layers are followed by instance normalization~\cite{ulyanov2017improved}. Similar to MUNIT~\cite{huang2018munit}, each residual block contains two adaptive instance normalization layers~\cite{huang2017arbitrary}. \textbf{(4)} For the discriminator $D$, we apply the multi-scale discriminator (PatchGAN) proposes by Wang \emph{et al.} \cite{wang2018high}. Moreover, we adopt the Leaky ReLU with slope 0.2 in both generator and discriminator. 
%In particular, we show detailed network architectures of $E^{e}_{.}$, $E^{f}_{.}$, and $G_{.}$ in TABLE \ref{tab:eye encoder}, TABLE \ref{tab:face encoder}, and TABLE \ref{tab:decoder}.

%\vspace{-1mm}
During the training, we adopt Adam optimizer~\cite{kingma2014adam} to optimize the generator and the discriminators. In addition, we set the initial learning rate to 0.0001, weight decay 0.0005, and exponential decay rates $(\beta_{1},\beta_{2})=(0,0.999)$. Following several typical image-to-image translations~\cite{liu2017unsupervised,huang2018munit, lee2018diverse}, we set hyper-parameters $\lambda_{face}=10$ for the face self-reconstruction loss. To encourage the model to focus on the eye region, we set a large weight of $\lambda_{eye}=10$ for the eye-area reconstruction loss. For the adversarial loss $ L_{adv}$, we employ the LSGAN objective proposed by Mao \emph{et al.} \cite{mao2017least}. Moreover, the gradient punishment strategy \cite{mescheder2018training} is also adopted to stabilize our model training procedure. 

\vspace{-.15in}
\subsection{Competitive Methods}
We compare the proposed method with several two-domain image-to-image translation frameworks, including CycleGAN~\cite{CycleGAN2017}, UNIT~\cite{liu2017unsupervised}, and MUNIT~\cite{huang2018munit}. These competitive methods usually treat the face with different attributes as two or multiple domains. By contrast, we view the face images as one domain with two codes, \emph{i.e.,} the face appearance code, and the eye-area attribute code. Besides, these three competitive methods manipulate the whole image to remove glasses, while our method only operates the eye area and other regions remain unchanged. To make a fair comparison, we train our method and \cite{CycleGAN2017, liu2017unsupervised, huang2018munit} under the same setting.

% To fairly evaluate the performance of our method with three competitive methods, we train the proposed method and these methods with the same training data and test on both CelebA and LFW.

% \textbf{CycleGAN~\cite{CycleGAN2017}.}
% CycleGAN combines adversarial loss and cycle consistency loss to train two generative adversarial networks for image-to-image translation.
 
% \textbf{UNIT~\cite{liu2017unsupervised}.}
% The UNIT model assumes that images of different domains could be mapped to the same latent representation, and the images generated by the model could be associated with input images of different domains by variational autoencoders~\cite{kingma2013auto}, respectively. 

% \textbf{MUNIT~\cite{huang2018munit}.} The MUNIT model is a multi-modal unsupervised image-to-image translation framework. It assumes that images from different domains could be decomposed into a shared content space and a style specific space. Therefore, an image could be translated to the target domain by recombining the content code of the image with a random style code in the target style space. 

\begin{figure*}[t]
\begin{center}
   \includegraphics[width=1.0\linewidth]{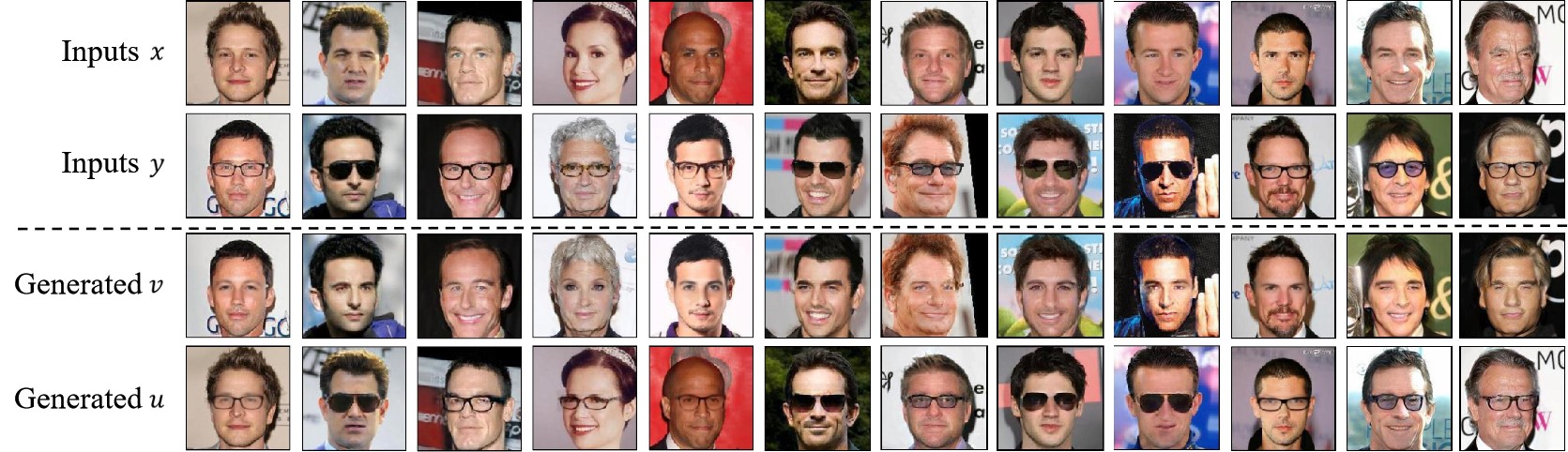}
 \end{center}
   \caption{Examples of our generated images by swapping eye-area attribute codes on the CelebA dataset.}
 \label{fig:swap}
 \end{figure*} 
\vspace{-.15in}
\subsection{Evaluations}
\textbf{Qualitative evaluation.}
We first qualitatively compare our method with three generative approaches above-mentioned.
As shown in Fig. \ref{fig:compare} and Fig. \ref{fig:compare_lfw}, we evaluate the generated results quality after eyeglasses removal from face images on the CelebA dataset and the LFW dataset, respectively. We re-implement the competitive methods, \emph{i.e.,} CycleGAN~\cite{CycleGAN2017}, UNIT~\cite{liu2017unsupervised}, and MUNIT~\cite{huang2018munit} by the open-source code. As shown in Fig. \ref{fig:compare} and Fig. \ref{fig:compare_lfw}, CycleGAN, UNIT, and MUNIT still have limitations in the manipulation of eyeglasses removal. These competitive methods are prone to generate blurry or over-smoothing results and remove glasses insufficiently. In comparison, our generated images are natural and realistic, suggesting that the proposed method is effective in removing eyeglasses from face images. In particular, as illustrated in Fig. \ref{fig:residual}, the manipulation of eyeglasses removal by three competitive methods also change rest regions outside the eye area, which dramatically reduces the quality of the generated image. By contrast, our method only manipulates the eye region while keeping the rest regions unchanged.

\begin{figure}[t]
\begin{center}
   \includegraphics[width=1.0\linewidth]{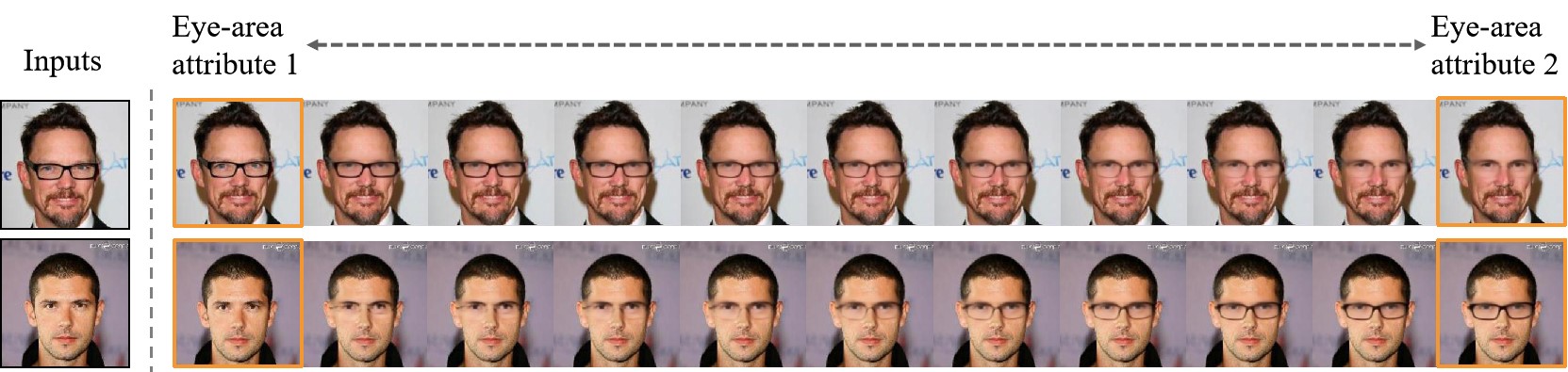}
 \end{center}
   \caption{\textbf{Linear interpolation.} Image generation results with linear interpolation between two eye-area attribute codes.}
 \label{fig:smooth}
 \end{figure}

We show example-guided eye-area attribute manipulation results in Fig. \ref{fig:swap}. We observe that our method achieves high-quality results in two inverse operations, \ie, removing glasses and wearing glasses. Moreover, the eye-area feature from the target image maintains faithfully. We further perform linear interpolation between two eye-area attribute codes to generate the corresponding face images, as shown in Fig. \ref{fig:smooth}. The linear interpolation results demonstrate that the eye-area attribute of face images change smoothly with latent codes.

\textbf{Quantitative evaluation.}
Here we report the results of quantitative evaluations based on FID and eLPIPS metrics aiming to measure the realism and the diversity of our generated face images. As shown in TABLE \ref{table:visual}, our method obtains the minimum FID and the maximum eLPIPS on the CelebA dataset, suggesting that generated face images by our method have a closet distribution of real face images. We proceed to perform our method on the LFW dataset. As shown in TABLE \ref{table:lfw}, our method also achieves the minimum FID and the maximum eLPIPS. The result indicates that the proposed method has scalability compared to three competitive methods. Moreover,
the quantitative evaluation verifies the authenticity of qualitative evaluation, and our method is significantly superior to three competitive methods on both realism and diversity.

\begin{table}[t]
\centering
%\small
\caption{\textbf{Quantitative results.} Comparison of FID (lower is better) and eLPIPS (higher is better) to evaluate realism and diversity of generated face images and the real data on the CelebA dataset.} 
{
\label{table:visual}
\setlength{\tabcolsep}{2pt}
\begin{tabular}{c|c|c|c|c}
\shline
\multirow{2}{*}{Method} & \multicolumn{2}{c|}{FID $\downarrow$} & \multicolumn{2}{c}{eLPIPS $\uparrow$} \\  
\cline{2-5}
& \ Wearing\ \  & Removal &  \ Wearing\ \  & Removal \\ 
\shline
Real data & 6.02 & 5.62 & - & - \\ 
\shline
CycleGAN \cite{CycleGAN2017} & 15.65 & 20.67 & - & -\\ 
UNIT \cite{liu2017unsupervised} & 18.80 & 18.86 & 0.114 & 0.074 \\ 
MUNIT \cite{huang2018munit} & 29.42 & 18.85 & 0.283 & 0.144 \\
\shline
Ours w/o $L^{face}_{recon}$ & 14.12 & 16.60 & 0.005 & 0.002 \\
Ours w/o $L^{eye}_{recon}$ & 12.79 & 16.50 & 0.384& 0.162\\ 
Ours w/o $L^{f}_{recon}$ & 14.42 & 15.68 & 0.432 &  0.208\\
Ours w/o $L_{cc}$ & 12.46 & 16.27 & 0.426 &0.234  \\ 
Ours (\emph{half}) & - & 15.87 & - & 0.010  \\ 
Ours w/ $L^{e}_{recon}$ & 12.96 & 16.33 & \textbf{0.435} & 0.220\\
\shline
Ours (\emph{full}) & \textbf{11.96} & \textbf{15.07} & 0.428 & \textbf{0.240}\\ 
\shline
\end{tabular}}
\end{table}

\begin{table}[t]
\centering
%\small
\caption{\textbf{Quantitative results.} Comparison of FID (lower is better) and eLPIPS (higher is better) to evaluate realism and diversity of generated face images and the real data on the LFW dataset.} 
{
\label{table:lfw}
\setlength{\tabcolsep}{2pt}
\begin{tabular}{c|c|c|c|c}
\shline
\multirow{2}{*}{Method} & \multicolumn{2}{c|}{FID $\downarrow$} & \multicolumn{2}{c}{eLPIPS $\uparrow$} \\  
\cline{2-5}
& \ Wearing\ \  & Removal &  \ Wearing\ \  & Removal \\ 
\shline
Real data & 24.17 & 6.96 & - & - \\ 
\shline
CycleGAN \cite{CycleGAN2017} & 40.37 & 23.33 & - & -\\ 
UNIT \cite{liu2017unsupervised} & 40.30 & 35.31 & 0.132 & 0.076 \\
MUNIT \cite{huang2018munit} & 51.74 & 42.83 & 0.201 & 0.141 \\
\shline
Ours~(\emph{full})  & \textbf{26.58} & \textbf{19.87} & \textbf{0.367} & \textbf{0.260}\\
\shline
\end{tabular}}
\end{table}

\vspace{-.15in}
\subsection{Ablation Study}
\label{ablation}

To study the contribution of each component in the proposed method, we perform several variants of ERGAN and evaluate them on the CelebA dataset. Specifically, we evaluate six variants, \emph{i.e.,} \textbf{(1)} Ours w/o $L^{face}_{recon}$: our method without the face self-reconstruction loss term. \textbf{(2)} Ours w/o $L^{eye}_{recon}$: our method without the eye-area reconstruction loss term. \textbf{(3)} Ours w/o $L^{f}_{recon}$: our method without the face appearance reconstruction loss term. \textbf{(4)} Ours w/o $L{cc}$: our method without the cycle consistency loss term. \textbf{(5)} Ours (\emph{half}): our method without the dual learning scheme and only implementing the task of eyeglasses removal. \textbf{(6)} Ours w/ $L^{e}_{recon}$: our method with the eye-area attribute reconstruction loss term.

\begin{figure}[t]
\begin{center}
   \includegraphics[width=1.0\linewidth]{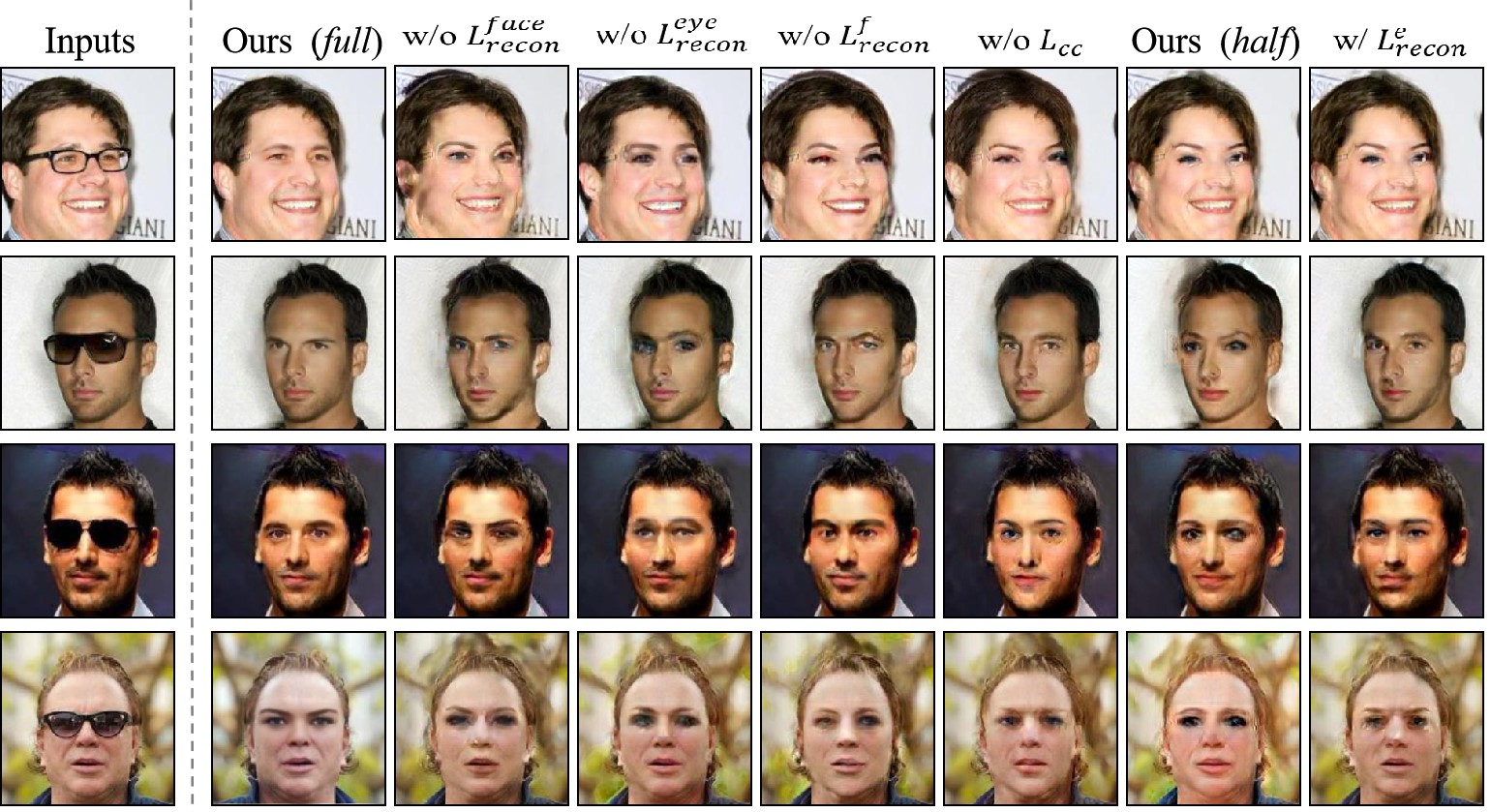}
 \end{center}
   \caption{\textbf{Qualitative ablation study.} Visualization results of six variants. }
 \label{fig:ablation}
 \vspace{-5mm}
 \end{figure}

We show the qualitative results of six variants in Fig.~\ref{fig:ablation}. Without using the face self-reconstruction loss, our method can still generate plausible results. However, some noise is introduced into the generated results, suggesting that the face self-reconstruction loss is a critical factor in keeping the rest regions except the eye area remains unchanged. Without adopting the face appearance reconstruction loss and the cycle consistency loss, our method generates slightly distorted results. The results can not preserve the consistency of the input image. Without employing the dual learning scheme and only implementing the task of eyeglasses removal, the model produces much lower quality results, demonstrating that the dual learning scheme is effective in learning features. We also evaluate the variant of our method with the eye-area attribute reconstruction loss. The results show that adopting the eye attribute reconstruction loss can not improve the quality of generated images. We suspect that this constraint is too strong, increasing the interferences in such a small region of the eye area. To achieve the best performance of our method, we remove the eye attribute reconstruction loss. 

We report the quantitative ablation study results of six variants in TABLE \ref{table:visual}. It can be observed that the full method obtains a lower FID score than six variants, suggesting that the proposed method can generate more realistic images on two tasks. For diversity, the scores drop dramatically without applying the face self-reconstruction loss, which indicates that this constraint is the key component to generate diversity outputs. In comparison to the variant which only implements the single task of eyeglasses removal, our method achieves lower FID scores and higher eLPIPS scores. It demonstrates that adopting the dual learning scheme can generate higher quality images in terms of realism and diversity. We observe that our method introduces the eye-area attribute reconstruction loss performs slightly better on glasses wearing in the diversity metric. The proposed method achieves the best performance on all the other indications, especially on eyeglasses removal in the realism metric.
\vspace{-.15in}
\subsection{Further Analysis and Discussions}
\begin{figure}[t]
\begin{center}
   \includegraphics[width=1.0\linewidth]{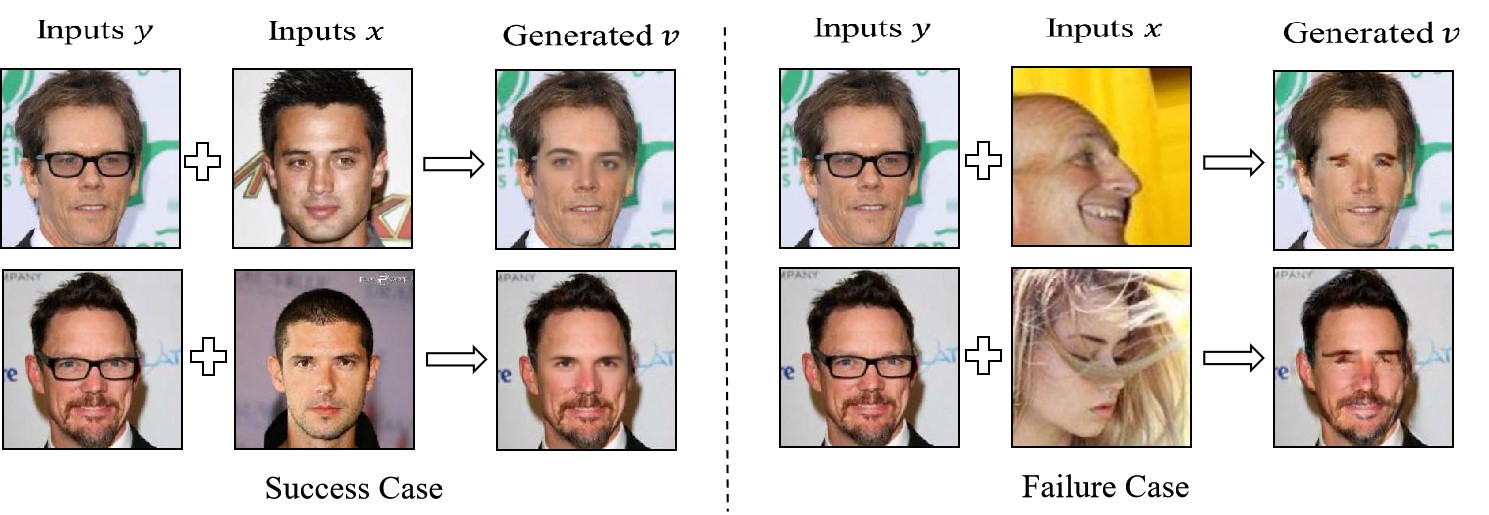}
 \end{center}
 \vspace{-2mm}
   \caption{The success and failure cases of eyeglasses removal applying our method.}
 \label{fig:failure}
\end{figure}

\textbf{Limitation.} We notice that the proposed method tends to produce low-fidelity results when the input face image pairs have large pose changes. As shown in Fig. \ref{fig:failure}, given two face images with a similar pose, our method could generate high-quality results. By contrast, the quality of generated images is degraded when the input images have significant pose changes (\emph{e.g.,} one front face and one profile face). There are two main reasons for the limitation. One is the limited amount of training data. Therefore, it is hard for our method to learn the two representations, \emph{i.e.}, the face appearance code and the eye-area attribute code, to cover all the variations introduced by poses. Another reason is that the profile face usually contains noises such as background and hairs, compromising the  eye-area attribute code learning. As a result, the generated face is hard to simulate the target eye.

\begin{table}[t]
\centering
%\small
\caption{\textbf{Quantitative face verification results on LFW.} Comparison of the face verification accuracy on $P1$, $P2$, and $P2_{r}$ to evaluate the effectiveness of eyeglasses removal. $P2$ contains face images with glasses which select from original images. $P2_{r}$ contains the same images with $P2$ while those images are manipulated with glasses removal adopting our method}.
\label{tab:face}
{
\setlength{\tabcolsep}{4.5pt}
\begin{tabular}{c|c}
\shline
   Face images & Accuracy \\
\hline
   Original test images \textbf{$P1$} & 97.80\% \\
   Selected images \textbf{$P2$} & 96.35\% \\
   Selected images after eyeglasses removal \textbf{$P2_{r}$} & 96.67\% \\
\shline
\end{tabular}}
%\vspace{+2mm}
\end{table}

\begin{table}[t]
\centering
%\small
\caption{\textbf{Quantitative face verification results on MeGlass.} Comparison of the face verification accuracy on $M1$, $M2$, and $M2_{r}$ to evaluate the effectiveness of eyeglasses removal. $M1$ contains face images without glasses. $M2$ only includes face images with glasses. $M2_{r}$ contains the same images with $M2$ while those images are manipulated with glasses removal adopting our method.}
\label{tab:meglass}
{
\setlength{\tabcolsep}{11pt}
\begin{tabular}{c|c}
\shline
   Face images & Accuracy \\
\hline
   Images without eyeglasses  \textbf{$M1$} & 96.34\% \\
   Images with eyeglasses \textbf{$M2$} & 94.00\% \\
   Images after eyeglasses removal \textbf{$M2_{r}$} & 94.66\% \\
\shline
\end{tabular}}
%\vspace{+2mm}
\end{table}

\begin{figure*}[t]
%\vspace{+2mm}
\begin{center}
   \includegraphics[width=0.90\linewidth]{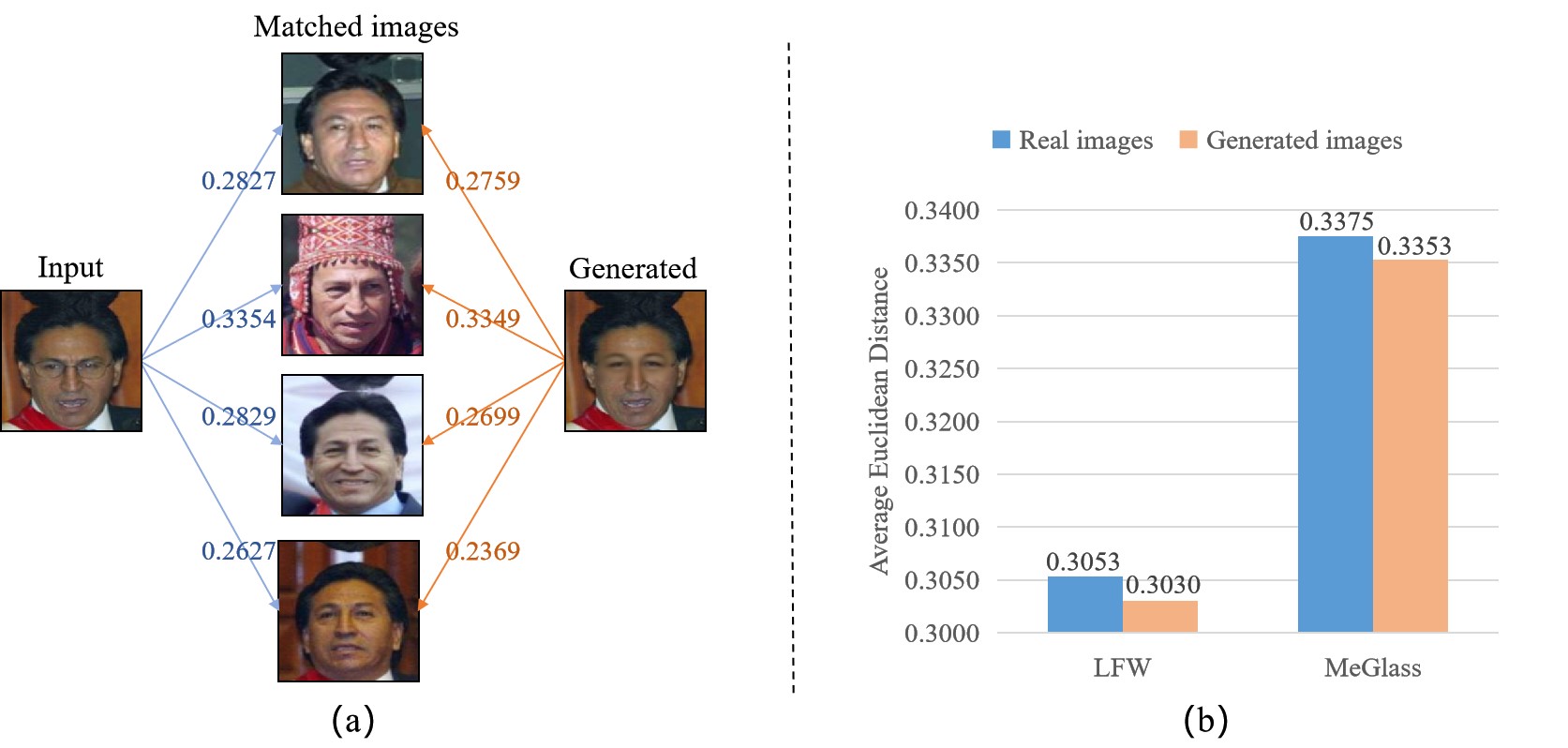}
 \end{center}
   \caption{Comparison of the Euclidean Distance (lower is better) to evaluate the effectiveness of eyeglasses removal by our method. (a) The left image is one input face image with eyeglasses. The middle images are other face images of the same identity. The right image is the generated image which is removed glasses by our method. The number on the line denotes the Euclidean distance between two images. (b) The  bar denotes the Average Euclidean Distance of the same person with and without eyeglasses, and the orange bar denotes the Average Euclidean Distance of the same person without glasses and removing glasses by ERGAN.}
 \label{fig:verification}
 \end{figure*}

\textbf{Face verification.} 
To further evaluate the performance of the proposed method, we first test whether it is beneficial to the face verification task. We apply the off-the-shelf face verification model FaceNet\footnote{\url{https://github.com/davidsandberg/facenet}} \cite{schroff2015facenet}.
Three test subsets of the LFW dataset are taken into consideration, \emph{i.e.,} original test images $P1$, selected images with eyeglasses $P2$, and selected images removed glasses by the proposed method $P2_r$.
$P1$ is the standard test set of the LFW dataset. Due to the limited number of face images with eyeglasses in LFW, we further sample a subset $P2$ from the original test set, which contains both matched pairs (with the same identity, one face image with glasses and another without glasses) and mismatched pairs (with different identities, both face images with glasses). The subset $P2_{r}$ contains the same images as $P2$, while those images are manipulated with glasses removal by our method. 

We report quantitative face verification results on LFW in TABLE \ref{tab:face}. Comparing the face verification results obtained on $P1$ and $P2$, we observe that the accuracy obtained on the test set $P2$ is lower than that obtained on $P1$. This result shows that the occlusion of glasses compromises the accuracy of the face verification model. However, this result does not indicate that the occlusion of glasses is the only factor that reduces accuracy. Other factors, including illumination, resolution, and viewpoint, still account for the reduced accuracy of face verification. Moreover, comparing the results between $P2$ and $P2_{r}$, we find that the performance of face verification improved after eyeglasses removal by the proposed method. Notably, it is reasonable to obtain a gain of 0.32\% for face verification. Our method can alleviate the impact of glasses occlusion on face verification. Although other factors such as illumination, resolution, and viewpoint exist and affect the accuracy of face verification, our method can improve the accuracy to a sensible extent. 

For face verification, we also perform experiments on three subsets of the MeGlass dataset, \ie, $M1$ that contains face images without glasses, $M2$ that contains face images with glasses, and $M2_{r}$ that includes the same samples as $M2$. $M2_{r}$ includes all images that have been manipulated with eyeglasses removal by our method. We report quantitative face verification results on MeGlass in TABLE \ref{tab:meglass}. We find that the face verification accuracy obtained on the test set $M2_{r}$ is higher than the accuracy on $M2$. The result is consistent with our statement that eyeglass removal could help face verification. Therefore, the proposed method could improve the accuracy of face verification on the large-scale dataset. 

We show the qualitative face verification results on LFW in Fig. \ref{fig:verification} (a). We observe that the Euclidean Distance decrease between matched images and the generated image by our method. Moreover, we present results of the Average  Euclidean Distance in Fig. \ref{fig:verification} (b). The  bar denotes the Average Euclidean Distance of the same person with and without eyeglasses, the orange bar denotes the Average Euclidean Distance of the same person without glasses and removing glasses by ERGAN. We observe that on both LFW and MeGlass, the Average Euclidean distance decrease between the same person without glasses and glasses removed by ERGAN. The result indicates that the proposed method could effectively alleviate the interference of glasses occlusion on face verification while preserving identification information well. Quantitative and qualitative results demonstrate that the proposed method is beneficial to the face verification task.

\textbf{Facial expression recognition.} Besides, we demonstrate that the proposed method benefits the facial expression recognition. To evaluate our method, we adopt a facial expression recognition algorithm\footnote{\url{https://github.com/opconty/keras-shufflenetV2}}, which effectively classifies the emotion into six adjectives (\ie, angry, scared, happy, sad, surprised, neutral). We perform experiments on the CelebA dataset. Notably, the CelebA data set only labels the attribute of smiling or not. Accordingly, we view that smiling represents the emotion of happy, and no-smiling represents other emotions. During the test, three sets are taken into consideration. $S1$ denotes that the set contains face images with attributes of smiling and no-glasses, $S2$ indicates that the set contains face images with attributes of smiling and glasses. $S2_{r}$ includes the same samples as $S2$, while all images are manipulated with eyeglasses removal by our method.

\begin{figure}[t]
%\vspace{+2mm}
\begin{center}
   \includegraphics[width=1.0\linewidth]{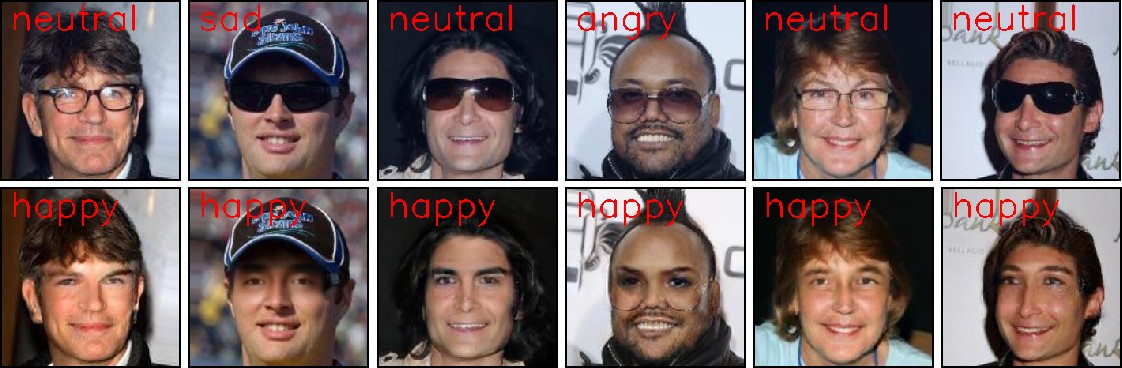}
 \end{center}
   \caption{\textbf{Qualitative facial expression recognition results.} The red word in the upper left corner of each image indicates the result of facial expression recognition. First row:  expression recognition results of original images. Second row:  expression recognition results of generated images by glasses removal. }
 \label{fig:expression}
 \end{figure}

We show qualitative results of facial expression recognition in Fig.~\ref{fig:expression}. We observe that face images with glasses in the first row are all misidentified as other expressions. By contrast, after the manipulation of eyeglasses removal by our method, all images are recognized as the happy expression in the second row. 

\begin{table}[t]
%\vspace{-2mm}
\centering
%\small
\caption{\textbf{Quantitative facial expression recognition results.} Comparison of the facial expression recognition accuracy on $S1$, $S2$, and $S2_{r}$ to evaluate the effectiveness of eyeglasses removal. $S1$ contains face images without glasses. $S2$ only includes face images with glasses. $S2_{r}$ contains the same images with $S2$ while those images are manipulated with glasses removal adopting our method.}

\label{tab:expression}
{
\setlength{\tabcolsep}{13pt}
\begin{tabular}{c|c}
%\small
\shline
   Face images & Accuracy \\
\hline
   Images without eyeglasses \textbf{$S1$} & 80.61\% \\
   Images with eyeglasses \textbf{$S2$} & 73.65\% \\
   Images after eyeglasses removal \textbf{$S2_{r}$} & 78.59\% \\
\shline
\end{tabular}}
\end{table}

We present quantitative results in TABLE~\ref{tab:expression}. Comparing the facial expression recognition results between $S1$ and $S2$, we find that the accuracy drops by 6.9\%. This result indicates that facial expression recognition is affected by the occlusion of glasses. Besides, comparing the results between $S2$ and $S2_{r}$, the accuracy of expression recognition gains by 4.9\%. 
The performance is close to the result on $S1$, which demonstrates the benefit of eyeglasses removal for facial expression recognition.

\section{Conclusion}
In this paper, we have proposed a GAN-based framework for eyeglasses removal in the wild. We adopt a dual learning scheme simultaneously to learn two inverse manipulations (removing glasses and wearing glasses), which enforces the model to generate high-quality results. Extensive qualitative and quantitative experiments demonstrate that our method outperforms previous state-of-the-art methods in terms of realism and diversity. Furthermore, we remark that the proposed method has the potential to be served as a pre-processing tool for other face-related tasks, \emph{e.g.,} face verification, and facial expression recognition. 
%\section*{Acknowledgment}

% \appendices

% % you can choose not to have a title for an appendix
% % if you want by leaving the argument blank
% \section{}

\ifCLASSOPTIONcaptionsoff
  \newpage
\fi



%\bibliographystyle{IEEEtran}

{\footnotesize
\bibliographystyle{IEEEtran}
\bibliography{egbib}
}


\end{document}